\documentclass[journal,usenames,dvipsnames]{IEEEtran}
\IEEEoverridecommandlockouts
\usepackage{cite}
\usepackage{amsmath,amssymb,amsfonts}
\usepackage{graphicx}
\usepackage{textcomp}

\usepackage{dsfont}
\usepackage{booktabs}
\usepackage[dvipsnames,table]{xcolor}
\usepackage{makecell}

\usepackage{algorithm,algorithmicx,listings}
\usepackage[noend]{algpseudocode}
\algnewcommand{\LineComment}[1]{{\Statex \vspace{1ex} \(\triangleright\) #1}}

\usepackage[
    breaklinks=true,
    colorlinks=true,
    linkcolor=blue,
    citecolor=blue,
    urlcolor=black
]{hyperref}
\usepackage{optidef}

\def\BibTeX{{\rm B\kern-.05em{\sc i\kern-.025em b}\kern-.08em
    T\kern-.1667em\lower.7ex\hbox{E}\kern-.125emX}}
\begin{document}

\title{Environment-Aware Probabilistic State Estimation of Continuum Robots from Planar Imaging\\
\author{
Aedan Mangan$^{1,2}$, Spencer Teetaert$^{2}$, Chengnan Shentu$^{2}$, Michael G. Brandel$^{3}$, Jeremy J. Heit$^{3}$, Alexander Norbash$^{4}$, Timothy D. Barfoot$^{2}$, Nikolay Atanasov$^{1}$, Tania K. Morimoto$^{1}$, Jessica Burgner-Kahrs$^{2}$
}  
\thanks{
$^{1}$Contextual Robotics Institute, University of California San Diego, USA. $^{2}$University of Toronto Robotics Institute, Canada. $^{3}$Departments of Radiology and Neurosurgery, Stanford University, USA. $^{4}$School of Medicine, University of Missouri-Kansas City, USA. AM was supported by the Merkin Graduate Fellows Program.
}
\thanks{This article has supplementary downloadable material (Supplementary Video S1) available, provided by the authors.
}
}

\maketitle

\begin{abstract}
Continuum robots are compliant along their entire length, making them well suited to navigating confined environments where contact with surrounding structures is often unavoidable. 
However, this compliance makes their modeling more complex compared to rigid robots, especially under contact, so their state must often be estimated from information available during operation.
In many applications, the environment is known in advance, for example from preoperative imaging in medicine or from design models in industrial inspection.
We present a probabilistic state estimator that represents the known environment as a signed distance field, informing the estimated shape.
Continuous in arclength and time, it estimates the full state from asynchronous measurements anywhere along the robot.
It also fuses planar images of body-fixed markers from a moving camera with commanded actuation, so the robot needs no embedded sensors.
In a simulated aortic arch and a hardware phantom, the estimator tracked the robot into the target branch, and with a single view, using the known environment lowered the tip error from 2.5\% to 1.6\% and from 20.0\% to 11.3\% of the robot length, respectively.
The continuous estimate of the full state and its uncertainty can support navigation and control of sensorless continuum robots.
\end{abstract}


\section{Introduction} \label{sec_intro}

Continuum robots are compliant along their entire length, which makes them well suited to navigate confined and tortuous environments, such as those in medical applications~\cite{burgner2015continuum,dupont2022continuum}.
In these environments, contact between the robot body and its surroundings is often unavoidable, and operators even use it to guide the robot tip~\cite{rao2024towards,mangan2026contactawareplanningcontrol}.
Contact also bends the robot body in ways that models predict poorly, and because contact along the body steers the tip and informs control, its entire shape must be estimated, not only its tip.

Embedded sensors such as fiber Bragg gratings and electromagnetic trackers supply shape information~\cite{al2020improved,ha2021robust} but add cost and size to slender, often single-use instruments~\cite{ramadani2022survey}.
External imaging, such as cameras, ultrasound, and fluoroscopy, observes the robot without requiring embedded sensors, but a single planar image leaves the out-of-plane shape ambiguous~\cite{ramadani2022survey,lu2023image}.
Additional views require more imaging hardware, which is not always available or practical, or moving the camera, so the views are no longer simultaneous, and in fluoroscopy both add radiation exposure~\cite{lobaton2013continuous,ramadani2022survey}.

Robotic systems also offer two sources of information that require no embedded sensors.
The controller knows the commanded actuation at every instant, and the geometry of the surroundings is often available beforehand, such as the anatomy segmented from preoperative imaging. 
Neither suffices alone, since mapping actuation to shape relies on the same models that fail under contact~\cite{orekhov2023lie,bajo2011kinematics}, and the environment constrains where the robot can be without revealing where it is.

\begin{figure}[t!]
\centerline{\includegraphics[width=0.9\linewidth]{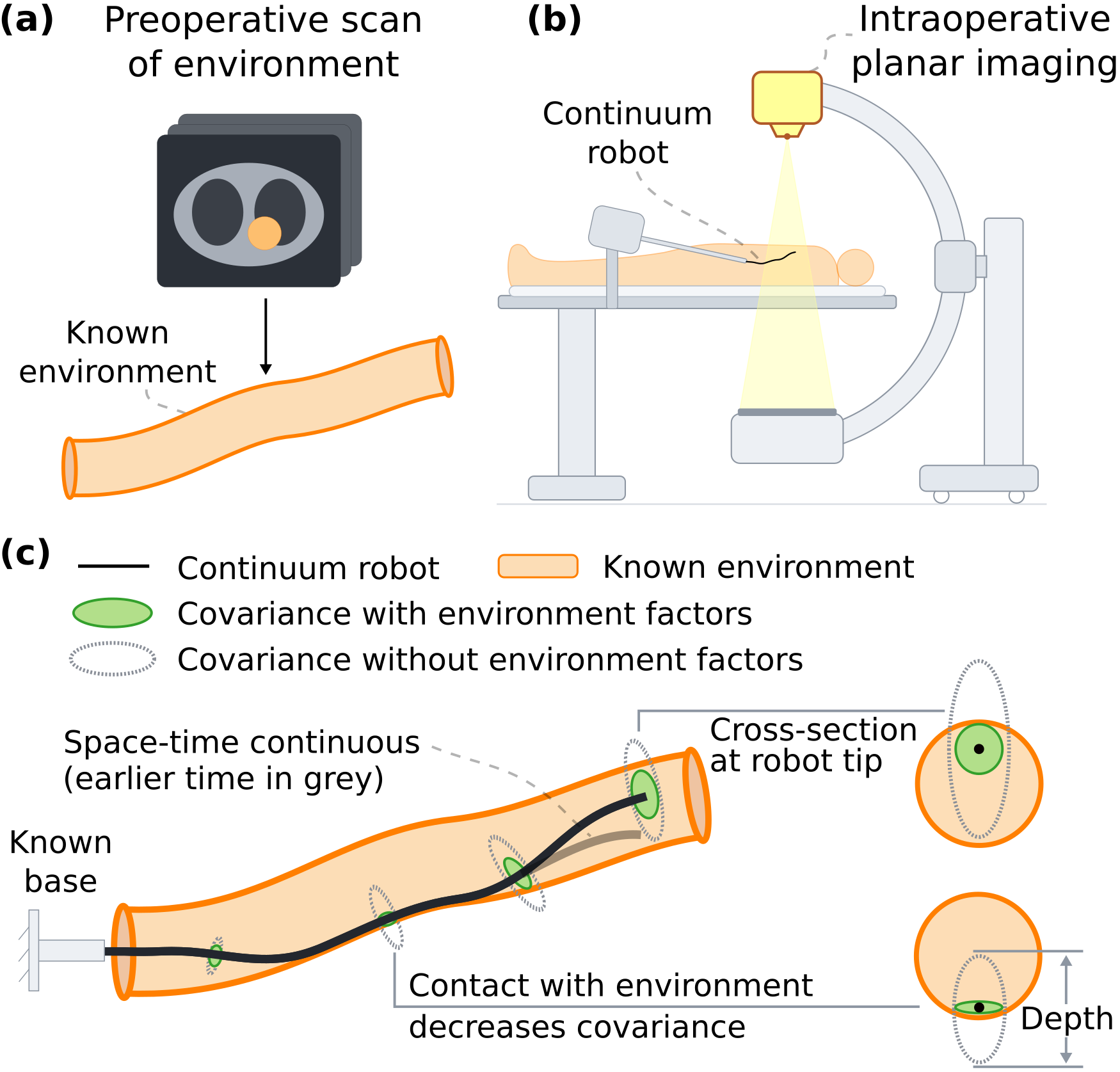}}
\vspace{-8pt}
\caption{Environment-aware probabilistic state estimation. 
(a)~A known environment from a preoperative scan (e.g., the aortic arch) is fused with 
(b)~intraoperative planar imaging and robot actuation to estimate 
(c)~the continuum robot's pose, strain, and velocity $\mathbf{x}(s,t)=\{\mathbf{T}_{wb},\boldsymbol{\varepsilon},\boldsymbol{\varpi}\}$, continuous in arclength $s$ and time $t$. 
Ellipses show covariance with (green) and without (dashed gray) environment factors. 
Accounting for the known environment shrinks the depth uncertainty that planar imaging alone cannot resolve.}
\label{fig_Purpose}
\end{figure}

Probabilistic estimation provides a principled way to fuse such information. 
Gaussian process estimators implemented via factor graphs yield continuous backbone estimates with uncertainty, and recent works extend them across both arclength and time~\cite{lilge2022continuum,ferguson2024unified,ferguson2026continuumactuation, teetaert2025stochastic,ferguson2026bayesian}. 
However, none of these estimators constrains the shape directly with known surrounding geometry, and they rely on intrinsic sensing, such as electromagnetic trackers, fiber Bragg gratings, or tendon tension sensors on the robot.
We hypothesize that a known environment improves the estimate because it bounds both the deflections that actuation models predict poorly and the depth that a single camera view leaves ambiguous.
We address both gaps (Fig.~\ref{fig_Purpose}) and evaluate the estimator in simulation and on hardware.
Our contributions are summarized as follows.
\begin{itemize}
    \item To our knowledge, the first state estimation approach in which contact with known environment geometry, anywhere along the body, informs the full shape estimate of an actuated continuum robot.
    \item A state estimator for multi-segment tendon-driven continuum robots that uses only extrinsic sensing and commanded actuation, so the robot needs no embedded sensors.
    It fuses planar images of body-fixed markers from a camera that may move between frames with a kinematic model of the commanded tendon displacements and insertion, so neither the camera nor the robot base needs to be stationary.
    \item The first demonstration of probabilistic estimation of a multi-segment continuum robot navigating a non-convex model of the aortic arch, recovering the full shape and its uncertainty continuously in arclength and time, in simulation and on hardware.
\end{itemize}

\section{Related Work} \label{sec_related_work}

\subsection{Continuum Robot Shape Estimation}
The shape of continuum robots has been sensed with external cameras, embedded fiber Bragg gratings, electromagnetic trackers, and the lengths of tendons and passive strings~\cite{ramadani2022survey,al2020improved,hoffmann2026reconstruction,lu2023image,orekhov2023lie}. 
Many of these methods fit a curve to each configuration independently, for example by matching image pixels to the backbone~\cite{hoffmann2026reconstruction,lu2023image} or by solving tendon and string length constraints over a modal curvature basis~\cite{orekhov2023lie}. 
Such fits can be viewed as maximum likelihood estimates that weight all measurements equally, and because each configuration is solved independently, they do not exploit consistency over time.

Maximum a posteriori (MAP) estimators instead combine measurements, weighted by their noise, with a prior over the robot state, and report the uncertainty of the result. 
They have been implemented with Kalman filters over curvature models~\cite{zhang2025stochastic}, Gaussian process regression along arclength~\cite{lilge2022continuum,ferguson2024unified}, and factor graphs over a discretized rod~\cite{ferguson2026continuumactuation,prakash2026multicontact}. 
Recent work extends these estimators across time with continuous-time motion priors~\cite{teetaert2025stochastic,teetaert2026tof} or with Cosserat rod dynamics~\cite{ferguson2026bayesian}. 
These estimators are driven by pose, position, strain, or range measurements, and none constrains the estimate directly with known surrounding geometry.

Contact with the environment is a large contributing factor to why the actual shape departs from the shape predicted by actuation and a model. 
Bajo and Simaan detect and localize contacts by comparing commanded and measured motion~\cite{bajo2011kinematics}, and shape-based force estimation recovers contact loads along the backbone~\cite{ferguson2024unified,prakash2026multicontact,ferguson2026bayesian}. 
When the environment is known, contact-aware kinematic models predict the shape by minimizing bending energy subject to actuation and obstacle constraints, and have been used to plan navigation~\cite{ashwin2021profile,rao2024towards,mangan2026contactawareplanningcontrol} but not to estimate the state from measurements. 

Actuation carries shape information that is most valuable when sensing is sparse~\cite{lilge2025incorporating,ferguson2026continuumactuation}. It has entered estimation as control inputs to a Gaussian process prior~\cite{lilge2025incorporating}, as boundary conditions in continuous arclength estimation~\cite{ferguson2024unified}, as uncertain tendon tensions in factor graphs~\cite{ferguson2026continuumactuation}, and as actuation readings that alone yield shape and forces~\cite{feliu2025actuation}. 
For tendon-driven robots, however, these estimators rely on measured tension and a mechanics model, even when the actuator prescribes tendon length~\cite{ferguson2024unified}. 
Tendon lengths have otherwise been used only in per-configuration shape fits~\cite{orekhov2023lie} and as hard constraints in planning~\cite{mangan2026contactawareplanningcontrol}, and not as uncertain constraints in a MAP estimate over arclength and time.

\subsection{Device Tracking in Endovascular Interventions}
Catheter and guidewire tracking in fluoroscopy is well studied~\cite{ramadani2022survey,chang2016robust}. 
Two or more views recover the three-dimensional shape through epipolar constraints~\cite{bender1999reconstruction,hoffmann2012semi,wagner2016interventional}, with the device found in each view by learning-based detection~\cite{hashemi2024use} or custom radiopaque markers~\cite{lawson2025fluoroscopic}, but they require biplane imaging or repeated C-arm repositioning and typically reconstruct each frame independently.
From a single view, depth must come from added assumptions. 
Band markers~\cite{hwang20183d} and constant curvature fits to a guidewire robot~\cite{ravigopal2023real} assume circular bending, which fails under environment contact. 

Preoperative vessel models can also supply depth from a single view.
Petkovi\'{c} et al. restricted a passive guidewire to the segmented vessels in each frame, so the vessel width limited the depth precision~\cite{petkovic2014realtime}.
Ambrosini et al. tracked a catheter tip over time along the vessel centerline, tolerating offsets within the vessel radius~\cite{ambrosini2016hidden}.
Both favor the centerline, which suits narrow vessels but not large ones such as the aortic arch, where the catheter is less constrained~\cite{beckers2026dynamic} and tends to lie against the wall~\cite{ambrosini2016hidden,mangan2026contactawareplanningcontrol}.
Neither uses the vessel geometry to estimate the full shape of an actuated robot.

\section{Problem Statement}  \label{sec_problem_statement}

\begin{figure}[t!]
\centerline{\includegraphics[width=0.9\linewidth]{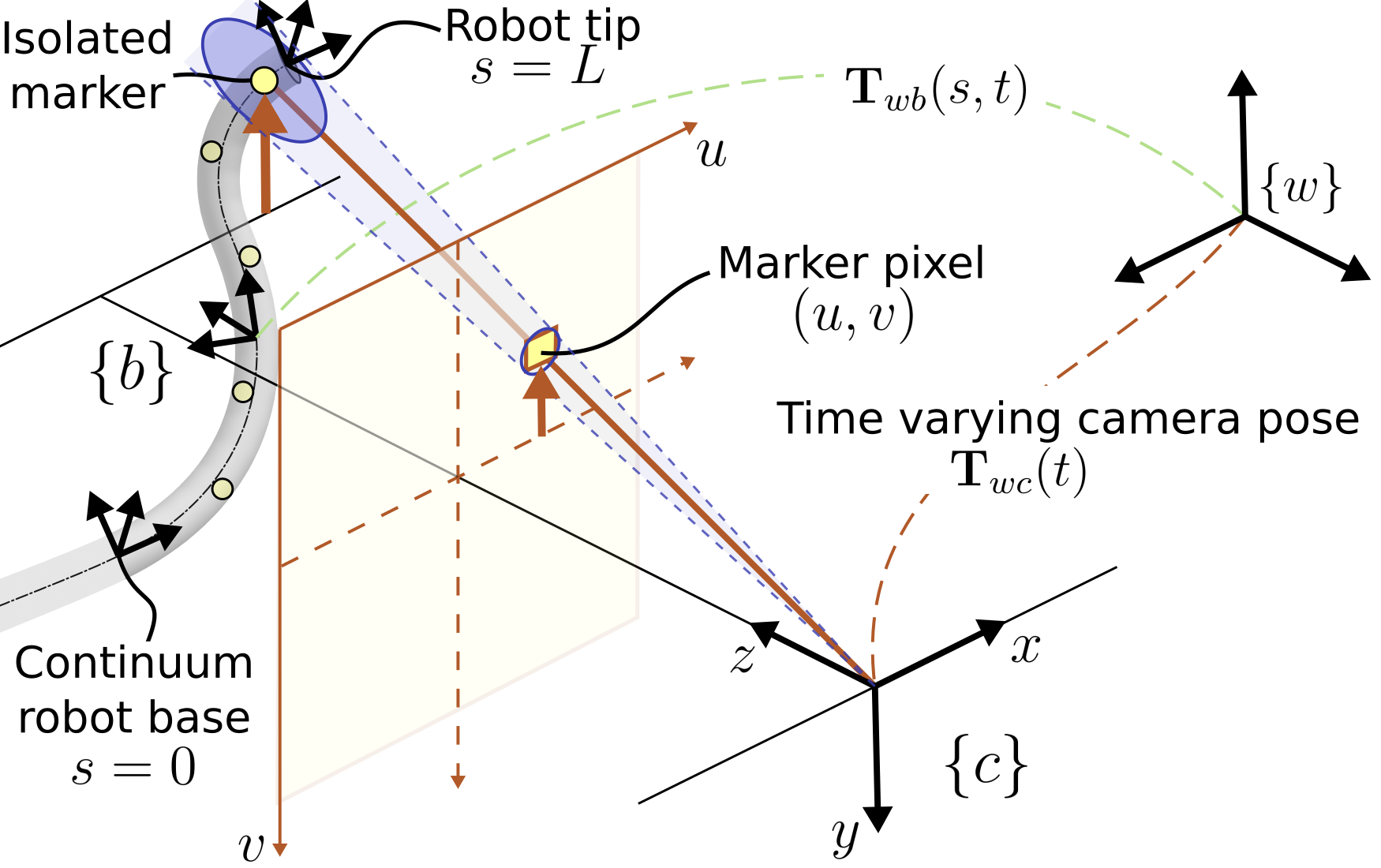}}
\vspace{-8pt}
\caption{The robot state is the pose $\mathbf{T}_{wb}(s,t)$ of the backbone frame $\{b\}$ at arclength $s$ and time $t$ relative to the world frame $\{w\}$. 
A marker at a known body-frame offset $\mathbf{r}_{b}$ is viewed by a camera with known, time-varying pose $\mathbf{T}_{cw}(t)$ and appears at pixel $(u,v)$ in the image. 
Observing body-fixed markers with an external camera allows the robot itself to remain sensorless.}
\label{fig_frame_convention}
\end{figure}

We consider the problem of estimating the state of a continuum robot over arclength and time as it moves through a known, static environment using imaging from an external camera. 
The robot body is modeled as a tube of radius $r$ around a continuous backbone of length $L$, parameterized by arclength $s\in[0,L]$ from the base (Fig.~\ref{fig_frame_convention}) and time $t$.
Its state is
\begin{equation} \label{eq_state}
\mathbf{x}(s,t)
=
\left\{
\mathbf{T}_{wb}(s,t),\,
\boldsymbol{\varepsilon}(s,t),\,
\boldsymbol{\varpi}(s,t)
\right\},
\end{equation}
with pose $\mathbf{T}_{wb}(s,t)\in SE(3)$ of the body frame $\{b\}$ of the cross section at arclength $s$, generalized strain $\boldsymbol{\varepsilon}(s,t)\in\mathbb{R}^{6}$, and generalized velocity $\boldsymbol{\varpi}(s,t)\in\mathbb{R}^{6}$, the latter two expressed in the body frame $\{b\}$.
The pose maps points expressed in $\{b\}$ to the world frame $\{w\}$,
\begin{gather}
\mathbf{T}_{wb}
=
\begin{bmatrix}
\mathbf{C}_{wb} & \mathbf{p}_{wb}\\
\mathbf{0}^{T} & 1
\end{bmatrix},
\\
\frac{\partial \mathbf{T}_{wb}}{\partial s}
=
\mathbf{T}_{wb}\,\boldsymbol{\varepsilon}^{\wedge},
\qquad
\frac{\partial \mathbf{T}_{wb}}{\partial t}
=
\mathbf{T}_{wb}\,\boldsymbol{\varpi}^{\wedge},
\label{eq_kinematics}
\end{gather}
where $\mathbf{C}_{wb}\in SO(3)$ and $\mathbf{p}_{wb}\in\mathbb{R}^{3}$ are the orientation and position of the cross section, respectively (Fig.~\ref{fig_frame_convention}). 
The operator $(\cdot)^{\wedge}$ maps $\boldsymbol{\xi}\in\mathbb{R}^{6}$, with rotational part $\boldsymbol{\phi}\in\mathbb{R}^{3}$ stacked above translational part $\boldsymbol{\rho}\in\mathbb{R}^{3}$, to the Lie algebra $\mathfrak{se}(3)$, and $(\cdot)^{\vee}$ is its inverse~\cite{barfoot2024state}.
We take a Bayesian approach, combining a prior over the continuous state with measurements to estimate its posterior.
Following~\cite{teetaert2025stochastic}, we model the state as a Gaussian process continuous in arclength and time, $\mathbf{x}(s,t)\sim\mathcal{GP}\bigl(\boldsymbol{\mu}(s,t),\boldsymbol{\Sigma}(s,t,s',t')\bigr)$, driven by white noise on the rates of change of the strain and velocity.
We approximate this prior with binary factors between adjacent nodes on a discrete grid in $(s,t)$, which compare the two nodes through their relative pose in its Lie algebra.
The prior also interpolates the state between nodes, so measurements do not need to coincide with them (Section~\ref{sec_prior}).
Three additional sources of information condition the prior:
\begin{itemize}
    \item \textbf{Camera.} We observe markers fixed to the robot body through a perspective projection with known intrinsics.
    We call any such projection a camera, including a fluoroscope, whose X-ray source and detector act as its center of projection and image plane.
    The camera pose $\mathbf{T}_{cw}(t)\in SE(3)$, which maps points from $\{w\}$ to the camera frame $\{c\}$, is known and may change between images, as when a C-arm rotates between views.
    Each marker lies at a known arclength $s$ and offset $\mathbf{r}_{b}\in\mathbb{R}^{3}$ in $\{b\}$, and each measurement $\tilde{\mathbf{z}}\in\mathbb{R}^{2}$ is its pixel location in one image.
    \item \textbf{Environment.} The workspace $\mathcal{W}\subset\mathbb{R}^{3}$ contains a known, static set of obstacles $\mathcal{O}\subset\mathcal{W}$ of any number and shape. 
    It is represented by the signed distance field $d:\mathbb{R}^{3}\to\mathbb{R}$, $d(\mathbf{p}) = \pm\min_{\mathbf{q}\in\partial\mathcal{O}}\|\mathbf{q}-\mathbf{p}\|$, which is negative inside $\mathcal{O}$, zero on its surface, and positive in free space.
    The robot body must stay clear of $\mathcal{O}$, $d(\mathbf{p}_{wb}(s,t))\geq r$, which we impose as a one-sided penalty at sampled arclengths and times.
    
    \item \textbf{Actuation.} The input $\mathbf{u}(t)$ consists of the commanded lengths $\tilde{\boldsymbol{\ell}}(t)\in\mathbb{R}^{M}$ of $M$ tendons, each routed at a known body-frame offset and terminating at a known arclength, and the commanded insertion, which sets the base pose $\tilde{\mathbf{T}}_{wb}(t)$ at $s=0$.
    Rather than model how $\mathbf{u}(t)$ drives the dynamics, we relate it to the state kinematically, requiring $\tilde{\boldsymbol{\ell}}(t)$ to match the lengths $\boldsymbol{\ell}(\mathbf{x})$ of the tendon paths along the estimated backbone and $\tilde{\mathbf{T}}_{wb}(t)$ to match $\mathbf{T}_{wb}(0,t)$.
\end{itemize}

Given these, we seek the maximum a posteriori estimate of the state at all nodes,
\begin{equation} \label{eq_objective}
\hat{\mathbf{x}}
=
\operatorname*{arg\,min}_{\mathbf{x}}
\left(
J_{\mathrm{prior}}(\mathbf{x})
+ J_{\mathrm{camera}}(\mathbf{x})
+ J_{\mathrm{env}}(\mathbf{x})
+ J_{\mathrm{input}}(\mathbf{x})
\right),
\end{equation}
and approximate the posterior by a Gaussian with mean $\hat{\mathbf{x}}$ and covariance equal to the inverse of the information matrix, the Gauss-Newton approximation of the Hessian of~\eqref{eq_objective}, evaluated at $\hat{\mathbf{x}}$~\cite{barfoot2024state}.
Each term sums costs of the form $\tfrac{1}{2}\mathbf{e}(\mathbf{x})^{T}\mathbf{R}^{-1}\mathbf{e}(\mathbf{x})$ over residuals $\mathbf{e}(\mathbf{x})$ with covariance $\mathbf{R}$, which for the camera term are reweighted to bound the influence of outliers.

\section{Methods}  \label{sec_methods}
This section defines the terms of~\eqref{eq_objective}, namely the space-time prior, a camera factor for body-fixed markers, an environment factor that keeps the body out of $\mathcal{O}$, and tendon and insertion factors for the commanded actuation.
Section~\ref{sec_jacobians} then gives the Jacobians that the solver requires.
\subsection{Space-Time Prior}  \label{sec_prior}

The state is estimated at nodes $\mathbf{x}_{n,k} = \mathbf{x}(s_{n},t_{k})$ at arclengths $0=s_{0}<\dots<s_{N-1}=L$, which do not need to be evenly spaced, and times $t_{0}<\dots<t_{K-1}$.
We adopt the space-time Gaussian process prior of~\cite{teetaert2025stochastic}, in which the kinematics in~\eqref{eq_kinematics} are driven by white noise on the rates of change of the strain and velocity in both arclength and time.
Each pair of nodes adjacent in time or in arclength is connected by a binary factor whose residual compares the later node with the earlier one propagated by a step of constant velocity or constant strain, and each factor is weighted by the covariance that the white noise accumulates over that step~\cite{teetaert2025stochastic}.
The prior cost $J_{\mathrm{prior}}$ sums these costs over all pairs of adjacent nodes.
A unary factor holds the full state of the base node near a measured state at every time step in~\cite{teetaert2025stochastic}.
We keep a unary factor on the base node only at the first time step and let the insertion factor (Section~\ref{sec_insertion_factor}) inform the rest.
The camera, environment, and insertion factors are evaluated at the arclength and time of each sample, which in general do not coincide with a node.
The pose there is interpolated with the prior from the (up to four) surrounding nodes~\cite{teetaert2025stochastic}, and the chain rule carries each Jacobian of Section~\ref{sec_jacobians} through this interpolation to those nodes.

\subsection{Camera Factor}  \label{sec_imaging_factor}

A marker at arclength $s$ with body-frame offset $\mathbf{r}_{b}$ has the predicted pixel location
\begin{equation} \label{eq_final_obs_model}
h\bigl(\mathbf{x}(s,t)\bigr)
=
\pi\!\left(\mathbf{K},\;
\mathbf{T}_{cw}(t)\,\mathbf{T}_{wb}(s,t)
\begin{bmatrix}
\mathbf{r}_{b}\\
1
\end{bmatrix}
\right),
\end{equation}
where $\mathbf{K}$ is the intrinsic calibration matrix and $\pi(\mathbf{K},\cdot)$ is the standard perspective projection, which divides a point $\mathbf{m}_{c} = [X\;\,Y\;\,Z]^{T}$ in the camera frame by its depth $Z$ along the optical axis and returns the first two entries of $\mathbf{K}[X/Z\;\;Y/Z\;\;1]^{T}$ as pixel coordinates.

Let $\tilde{\mathbf{z}} = \begin{bmatrix}\tilde{u} & \tilde{v}\end{bmatrix}^{T}$ denote the measured pixel coordinates of a marker, with known association to its offset $\mathbf{r}_{b}$ and arclength $s$.
The residual is
\begin{equation} \label{eq_marker_residual}
\mathbf{e}_{\mathrm{marker}}
=
\mathbf{D}
\left(
\tilde{\mathbf{z}}
-
h\bigl(\mathbf{x}(s,t)\bigr)
\right)
\in\mathbb{R}^{2},
\end{equation}
where $\mathbf{D} = \operatorname{diag}(1/n_{u},\,1/n_{v})$ normalizes by the image width $n_{u}$ and height $n_{v}$ in pixels.
We model the residual as zero-mean Gaussian with covariance $\mathbf{R}_{\mathrm{marker}}$, a tuning parameter expressed in normalized image coordinates.
The weighted squared residual is $\nu = \mathbf{e}_{\mathrm{marker}}^{T}\mathbf{R}_{\mathrm{marker}}^{-1}\mathbf{e}_{\mathrm{marker}}$.
To bound the influence of outliers, each measurement contributes a Cauchy cost of scale $\gamma$, following~\cite{teetaert2026tof, barfoot2024state},
\begin{equation}
J_{\mathrm{camera}}
=
\sum_{\text{measurements}}
\frac{\gamma^{2}}{2}
\ln\!\left(1 + \frac{\nu}{\gamma^{2}}\right),
\end{equation}
which approaches the quadratic cost $\tfrac{1}{2}\nu$ for small residuals and down-weights a measurement by $1/(1+\nu/\gamma^{2})$, which is one half at a residual of $\gamma$ standard deviations.
Each Levenberg-Marquardt iteration handles this cost by iteratively reweighted least squares~\cite{barfoot2024state}, linearizing the marker residual with $\mathbf{R}_{\mathrm{marker}}$ divided by this weight, evaluated at the current estimate.
This weighted quadratic matches the gradient of the Cauchy cost at the current estimate, and each step is accepted only if it lowers the Cauchy cost itself.
The posterior covariance uses the weights at $\hat{\mathbf{x}}$, so a down-weighted marker contributes less information.
A marker predicted less than $1$~mm in front of the camera is assigned a constant residual with zero Jacobian so that it does not steer the estimate.



\subsection{Environment Factor}  \label{sec_environment_factor}

\begin{figure}[t!]
\centerline{\includegraphics[width=0.9\linewidth]{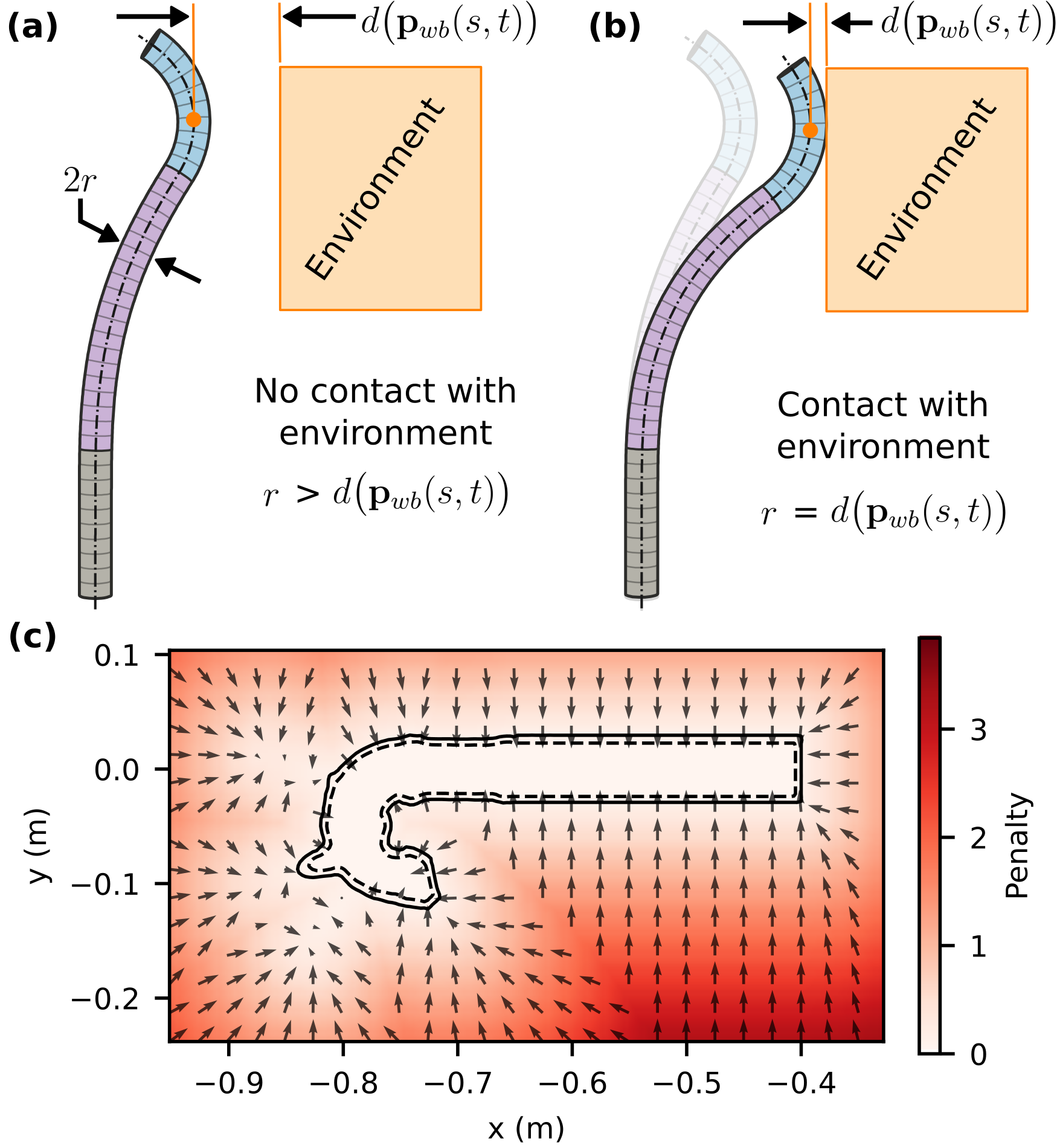}}
\vspace{-8pt}
\caption{
Environment factor based on the signed distance $d\bigl(\mathbf{p}_{wb}(s,t)\bigr)$ from a backbone point (orange) to the known environment $\mathcal{O}$, for a robot of radius $r$.
(a)~Without contact ($d > r$), the factor adds no cost.
(b)~In contact, $d = r$, and penetration ($d < r$) adds the residual $r-d$, pushing the backbone away from $\mathcal{O}$.
(c)~Penalty on one slice of the aortic arch scan, zero inside the dashed line offset by $r$ from the vessel wall (solid). 
The factor keeps the body out of $\mathcal{O}$ but leaves it free elsewhere.
}
\label{fig_env_factor}
\end{figure}

The signed distance field $d$ of $\mathcal{O}$ and its gradient $\nabla d$ are precomputed on a regular grid of cubic voxels in the world frame (Fig.~\ref{fig_env_factor}).
The constraint $d\bigl(\mathbf{p}_{wb}(s,t)\bigr)\geq r$ of Section~\ref{sec_problem_statement} is enforced softly, as a one-sided penalty on the penetration depth,
\begin{equation} \label{eq_env_residual}
e_{\mathrm{env}}
=
\max\Bigl(0,\;
r - d\bigl(\mathbf{p}_{wb}(s,t)\bigr)\Bigr)
\in\mathbb{R}.
\end{equation}
The residual is zero while the body is clear of $\mathcal{O}$ and grows linearly with the penetration, so the factor only shapes the estimate where the robot would otherwise intersect the environment (Fig.~\ref{fig_env_factor}).
A point outside the grid is treated as clear.

Because the factor is not tied to a sensor reading,~\eqref{eq_env_residual} is evaluated at a set of $\{s,t\}_i$, taken every $\Delta t_{\mathrm{env}}$ seconds at $N_{\mathrm{env}}$ arclengths evenly spaced from the base to the tip.
With standard deviation $\sigma_{\mathrm{env}}$, the cost is
\begin{equation} \label{eq_env_cost}
J_{\mathrm{env}}
=
\frac{1}{2}
\sum_{\text{samples}}
\frac{e_{\mathrm{env}}^{2}}{\sigma_{\mathrm{env}}^{2}} .
\end{equation}

\subsection{Tendon Factor}  \label{sec_tendon_factor}

\begin{figure}[t!]
\centerline{\includegraphics[width=0.75\linewidth]{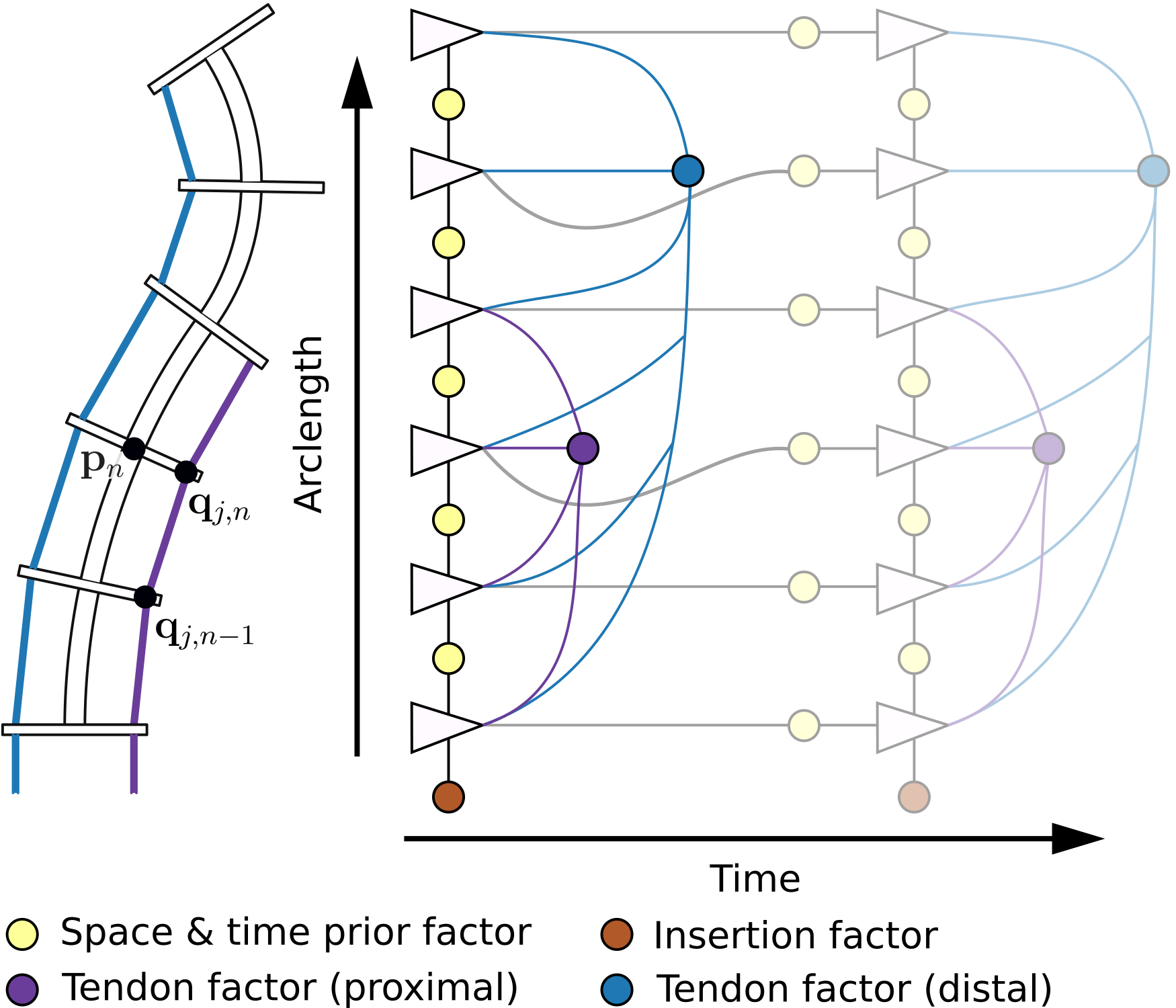}}
\vspace{-8pt}
\caption{
Factor graph of a two-segment tendon-driven continuum robot, with camera and environment factors omitted.
(Left)~Each tendon length is a polyline through routing points $\mathbf{q}_{j,n}$ offset from the backbone $\mathbf{p}_{n}$, for proximal (purple) and distal (blue) tendons.
(Right)~State nodes (triangles) at two time steps, linked by space-time prior, insertion, and tendon factors.
Each tendon length informs every node the tendon spans.
}
\label{fig_tendon_factor}
\end{figure}

Unlike the tension-based estimators of Section~\ref{sec_related_work}, we formulate a purely kinematic factor that needs only the commanded tendon lengths.
The input term $J_{\mathrm{input}} = J_{\mathrm{tendon}} + J_{\mathrm{base}}$ relates $\mathbf{u}(t)$ to the state through this tendon factor and the insertion factor of Section~\ref{sec_insertion_factor}.

We actuate the robot with $M$ tendons routed parallel to the backbone.
Tendon $j$ passes through each cross section at the constant body-frame offset $\mathbf{r}_{j} = [r_{t}\cos\theta_{j}\;\; r_{t}\sin\theta_{j}\;\; 0]^{T}$, with routing radius $r_{t}$ and angular position $\theta_{j}$ around the backbone.
At one time step, node $n=0,\dots,N-1$ has orientation $\mathbf{C}_{n}$ and position $\mathbf{p}_{n}$, so tendon $j$ passes node $n$ at the routing point $\mathbf{q}_{j,n} = \mathbf{C}_{n}\,\mathbf{r}_{j} + \mathbf{p}_{n}$ (Fig.~\ref{fig_tendon_factor}).
We approximate the tendon length by the polyline through its routing points,
\begin{equation} \label{eq_tendon_length}
\ell_{j}(\mathbf{x})
=
\sum_{n=1}^{N_{j}-1}
\bigl\lVert \mathbf{q}_{j,n} - \mathbf{q}_{j,n-1} \bigr\rVert ,
\end{equation}
where $N_{j}\leq N$ is the number of nodes the tendon spans from the base.

The value of $N_{j}$ distinguishes the segments of a multi-segment robot.
Each tendon terminates at the end of the segment it actuates, so a node is placed at every segment end and $N_{j}$ counts the nodes up to and including it.
In our two-segment design (Fig.~\ref{fig_tendon_factor}), the proximal tendons terminate at the node at the end of the proximal segment, while the distal tendons continue to the tip node, where $N_{j}=N$.
The same construction applies to any number of segments.

Let $\tilde{\boldsymbol{\ell}}\in\mathbb{R}^{M}$ denote the tendon lengths commanded at time $t$.
The residual stacks one entry per tendon,
\begin{equation} \label{eq_tendon_residual}
\mathbf{e}_{\mathrm{tendon}}
=
\boldsymbol{\ell}(\mathbf{x}) - \tilde{\boldsymbol{\ell}}
\in\mathbb{R}^{M}.
\end{equation}
A tendon length constrains every node along its path at once.
The factor is therefore attached directly to the node poses of the time step nearest to $t$, and not to an interpolated state.
The cost is
\begin{equation} \label{eq_tendon_cost}
J_{\mathrm{tendon}}
=
\frac{1}{2}
\sum_{\text{measurements}}
\mathbf{e}_{\mathrm{tendon}}^{T}\,\mathbf{R}_{\mathrm{tendon}}^{-1}\,\mathbf{e}_{\mathrm{tendon}} ,
\end{equation}
with diagonal covariance $\mathbf{R}_{\mathrm{tendon}}$.
When the nodes are spaced more widely than the spacer disks, the polyline is shorter than the tendon path, and the commanded lengths neglect tendon stretch and slack, so $\mathbf{R}_{\mathrm{tendon}}$ is set large to cover both.

\subsection{Insertion Factor}  \label{sec_insertion_factor}

The second part of $J_{\mathrm{input}}$ carries the insertion, which moves the base of the robot during operation. 
The base pose $\tilde{\mathbf{T}}_{wb}(t)$ is obtained from the commanded insertion and enters the estimator as a pose measurement at $s=0$.
We use the pose measurement factor of~\cite{teetaert2025stochastic}, written here for our pose convention, with residual and cost
\begin{gather}
\mathbf{e}_{\mathrm{base}}
=
\ln\!\left(
\mathbf{T}_{wb}(0,t)^{-1}\,\tilde{\mathbf{T}}_{wb}(t)
\right)^{\vee}
\in\mathbb{R}^{6}, \label{eq_base_residual}
\\
J_{\mathrm{base}}
=
\frac{1}{2}
\sum_{\text{measurements}}
\mathbf{e}_{\mathrm{base}}^{T}\,\mathbf{R}_{\mathrm{base}}^{-1}\,\mathbf{e}_{\mathrm{base}} , \label{eq_base_cost}
\end{gather}
where $\mathbf{R}_{\mathrm{base}}$ is the covariance of the commanded pose.

\subsection{Optimization and Jacobians}  \label{sec_jacobians}

We minimize~\eqref{eq_objective} with the Levenberg-Marquardt algorithm, which at each iteration linearizes every residual about the current estimate and therefore needs its Jacobian with respect to a small change in the state~\cite{barfoot2024state}.
Unlike the strain and velocity, which are vectors, the pose lies on $SE(3)$, so we perturb it by a small motion $\delta\boldsymbol{\xi}\in\mathbb{R}^{6}$, ordered like $\boldsymbol{\xi}$ and expressed in $\{b\}$, as $\mathbf{T}_{wb}\exp(\delta\boldsymbol{\xi}^{\wedge})$.
The Jacobian $\mathbf{H}$ of a residual with respect to this pose maps $\delta\boldsymbol{\xi}$ to the first-order change in the residual, $\delta\mathbf{e} = \mathbf{H}\,\delta\boldsymbol{\xi}$~\cite{boumal2023intromanifolds, barfoot2024state}.
None of the camera, environment, tendon, or insertion residuals depend on the strain or velocity, so only their pose Jacobians are nonzero.

The camera, environment, and tendon residuals depend on a pose through the world-frame position $\mathbf{y} = \mathbf{C}_{wb}\,\mathbf{r} + \mathbf{p}_{wb}$ of a point at a fixed body-frame offset $\mathbf{r}$, namely a marker, a tendon routing point, or the backbone itself ($\mathbf{r}=\mathbf{0}$).
Since $\exp(\delta\boldsymbol{\xi}^{\wedge})\approx\mathbf{I}+\delta\boldsymbol{\xi}^{\wedge}$, the perturbation moves this point by
\begin{equation} \label{eq_point_perturbation}
\delta\mathbf{y}
=
\mathbf{C}_{wb}
\begin{bmatrix}
-\mathbf{r}^{\wedge} & \mathbf{I}_{3}
\end{bmatrix}
\delta\boldsymbol{\xi},
\end{equation}
so each of these Jacobians is the derivative of the residual with respect to $\mathbf{y}$ multiplied by this matrix.

\subsubsection*{Camera}
With $\mathbf{C}_{cw}$ the rotation of $\mathbf{T}_{cw}$ and $\mathbf{H}_{\pi} = \partial\pi(\mathbf{K},\mathbf{m}_{c})/\partial\mathbf{m}_{c}$ the Jacobian of the perspective projection, the chain rule through~\eqref{eq_final_obs_model} and~\eqref{eq_point_perturbation} gives
\begin{equation}
\mathbf{H}_{\mathrm{marker}}
=
-\,\mathbf{D}\,\mathbf{H}_{\pi}\,\mathbf{C}_{cw}\,\mathbf{C}_{wb}
\begin{bmatrix}
-\mathbf{r}_{b}^{\wedge} & \mathbf{I}_{3}
\end{bmatrix}
\in\mathbb{R}^{2\times6}.
\end{equation}

\subsubsection*{Environment}
The environment residual is evaluated on the backbone ($\mathbf{r}=\mathbf{0}$), so while the body penetrates $\mathcal{O}$ ($d<r$), its Jacobian is $\mathbf{H}_{\mathrm{env}} = \begin{bmatrix}\mathbf{0}_{1\times3} & -\nabla d(\mathbf{p}_{wb})^{T} \mathbf{C}_{wb}\end{bmatrix}$, and it is zero otherwise.

\subsubsection*{Tendon}
Let $\mathbf{d}_{j,n}$ be the unit direction of the polyline segment $\mathbf{q}_{j,n} - \mathbf{q}_{j,n-1}$ for $n = 1,\dots,N_{j}-1$, with $\mathbf{d}_{j,0} = \mathbf{d}_{j,N_{j}} = \mathbf{0}$.
Each routing point enters~\eqref{eq_tendon_length} through the segments before and after it, so by~\eqref{eq_point_perturbation} the Jacobian of tendon $j$ with respect to the pose of node $n<N_{j}$ is
\begin{equation}
\frac{\partial e_{\mathrm{tendon},j}}{\partial \delta\boldsymbol{\xi}_{n}}
=
\bigl(\mathbf{d}_{j,n} - \mathbf{d}_{j,n+1}\bigr)^{T}
\mathbf{C}_{n}
\begin{bmatrix}
-\mathbf{r}_{j}^{\wedge} & \mathbf{I}_{3}
\end{bmatrix},
\end{equation}
and zero for $n \geq N_{j}$.
A segment of zero length has no direction and contributes no gradient.

\subsubsection*{Insertion}
The insertion residual~\eqref{eq_base_residual} acts on the base pose directly, and its Jacobian is $\mathbf{H}_{\mathrm{base}} = -\boldsymbol{\mathcal{J}}_{r}^{-1}(\mathbf{e}_{\mathrm{base}})\operatorname{Ad}(\tilde{\mathbf{T}}_{wb}^{-1}\mathbf{T}_{wb})$, with $\boldsymbol{\mathcal{J}}_{r}$ the right Jacobian and $\operatorname{Ad}(\cdot)$ the adjoint of $SE(3)$~\cite{barfoot2024state}.

\section{Experiments and Results} \label{sec_experiments}

We evaluated the estimator in simulation in free space, in contact with a simple environment, and in an aortic arch, and on hardware in an aortic arch phantom (see Supplementary Video S1).
We chose the aortic arch as a representative clinical example, since a catheter must traverse it during mechanical thrombectomy for acute ischemic stroke treatment, and physicians rely on contact between the device body and the vessel wall~\cite{mangan2025serially,ambrosini2016hidden,mangan2026contactawareplanningcontrol}.
In free space, the camera factor informed the position of the robot, while its uncertainty remained largest along the optical axis, which moved with each view.
The remaining experiments added the actuation and environment factors as the robot was commanded into contact, and showed that using the known environment reduced this depth uncertainty wherever the robot met it and lowered the tip error in the aortic arch, most of all when imaging was sparse.

\subsection{Implementation and Metrics} \label{sec_exp_metrics}

The estimator was implemented in C++ on the framework of~\cite{teetaert2025stochastic} using GTSAM~\cite{dellaert2012factor} and is released open source at (link provided upon acceptance). 
Every trial was solved as a batch with a Levenberg-Marquardt solver on an Intel i7-9750H CPU, with $N=6$ nodes in arclength and time steps of $0.1$~s in simulation and $0.4$~s on hardware.
In every trial, distinct colors for the markers gave the data association, and a Cauchy loss of scale $\gamma = 3$ on the camera factor limited the effect of misdetections.
The benchtop aortic arch phantom was a type~III aortic arch~\cite{MARROCCOTRISCHITTA20171584} scaled to twice the human size to match the size of the tendon-driven robot. 
It was scanned with a FARO Edge laser scanner. The resulting STL mesh served both as the simulated environment and, converted to a signed distance field, as the known environment of the estimator in simulation and on hardware.
Following~\cite{teetaert2025stochastic}, the estimate was evaluated at the tip, queried at the time of each ground truth sample.
We report the root mean square error (RMSE) of the tip position, normalized by the robot length $L$.
Consistency was measured by the average normalized estimation error squared (NEES) of the tip position, $\frac{1}{K}\sum_{k=1}^{K}\mathbf{e}_{k}^{T}\boldsymbol{\Sigma}_{k}^{-1}\mathbf{e}_{k}$ over the $K$ ground truth samples, with position error $\mathbf{e}_{k}$ and estimated covariance $\boldsymbol{\Sigma}_{k}$.
Its expected value is three for a consistent estimator, and a larger value indicates an overconfident one.

\subsection{Simulation} \label{sec_exp_sim}

\begin{figure}[t!]
\centerline{\includegraphics[width=0.8\linewidth]{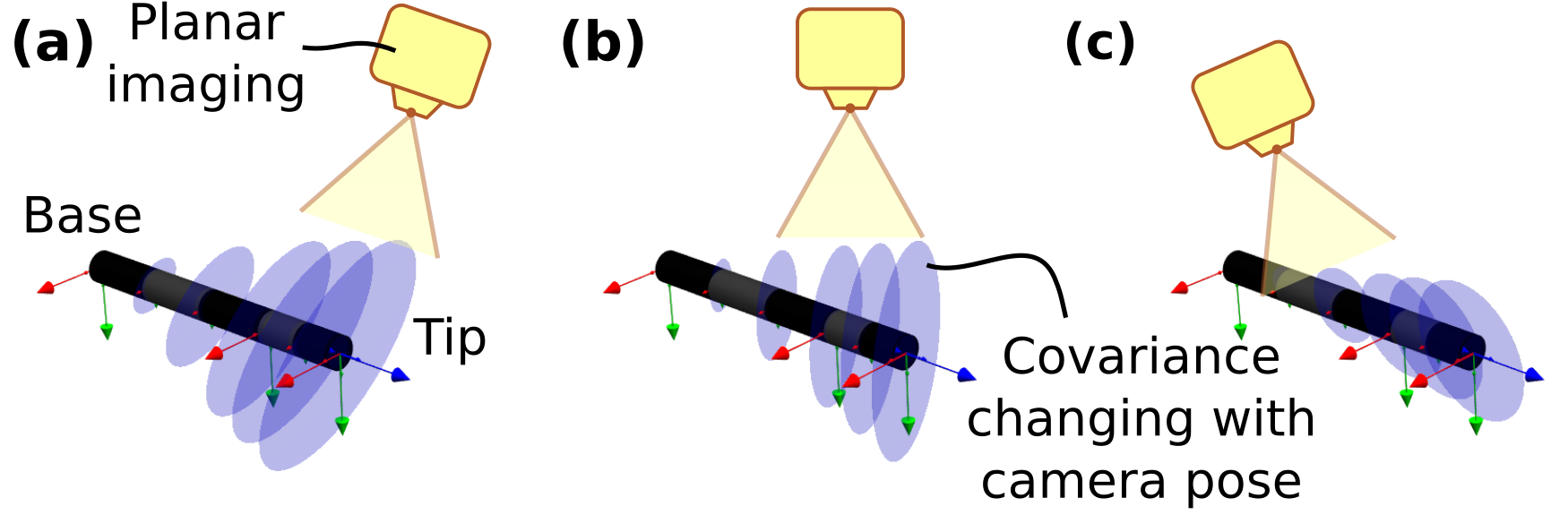}}
\vspace{-8pt}
\caption{
Illustration of the estimated covariance (blue) for three camera poses (see Supplementary Video S1).
A single view cannot resolve depth, so the covariance stretches along the optical axis and follows the camera pose.
}
\label{fig_exp_free_camera}
\end{figure}

We simulated a two-segment tendon-driven robot ($L=195$~mm, radius $5$~mm, four tendons per segment) in MuJoCo~\cite{shentu2026rigid}.
A camera on a Franka Panda robot arm rendered the seven markers on the robot from the frontal, left anterior oblique (LAO), and right anterior oblique (RAO) views, and the simulator provided the true pose along the robot (Fig.~\ref{fig_exp_free_camera}).

\subsubsection{Free Space}
With the robot in free space, the camera factor recovered its position in the image plane, while its uncertainty remained elongated along the optical axis, since a single view constrains each marker only up to its ray (Fig.~\ref{fig_exp_free_camera}).
As the camera moved between the frontal, LAO, and RAO views, this direction of largest uncertainty followed the optical axis, and combining the views reduced the largest $3\sigma$ tip uncertainty from $42$~mm with the frontal view alone to $26$~mm.

\begin{figure}
\centerline{\includegraphics[width=1.0\linewidth]{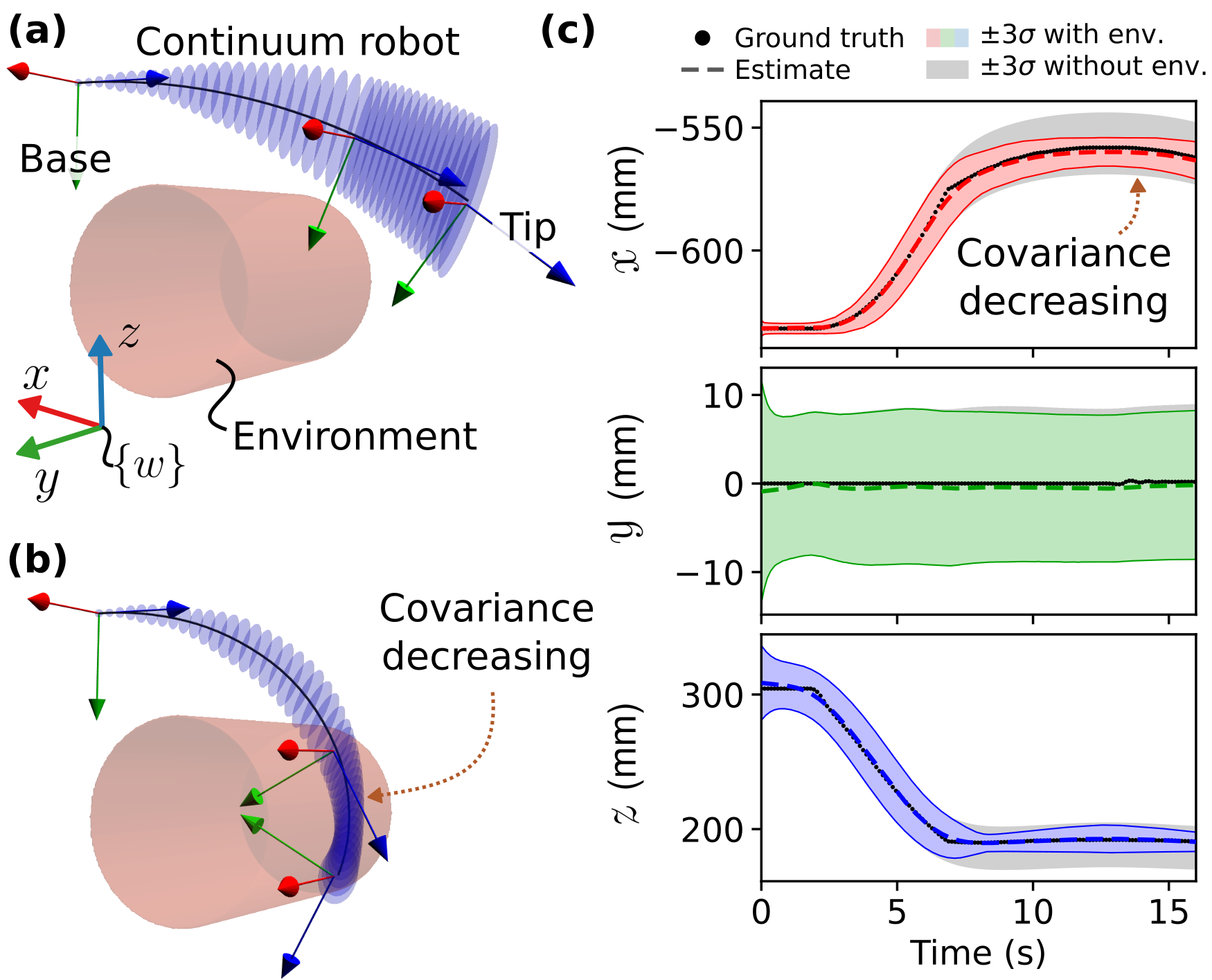}}
\vspace{-8pt}
\caption{
Continuum robot commanded into a known cylinder, imaged from the frontal view only.
(a)~Before contact ($t=4$~s) and (b)~in contact ($t=14$~s), with the estimated backbone (black), ground truth frames, and covariance ellipsoids (blue).
(c)~Estimated tip position (dashed) in $x$, $y$, and $z$ of $\{w\}$ with $\pm3\sigma$ bounds with (colored) and without (gray) the environment factor, and ground truth (black).
Contact with the known environment tightened the estimate.
}
\label{fig_exp_simple_env}
\end{figure}

\subsubsection{Contact with a Simple Environment}
To isolate the effect of the environment, the camera held the frontal view throughout, so only the actuation and environment factors constrained the depth along its optical axis, the $z$ axis of $\{w\}$.
The robot was then commanded into a cylindrical obstacle, with the commanded tendon lengths entering through the tendon factor.
While the robot was clear of the obstacle, the environment factor added no cost (Fig.~\ref{fig_exp_simple_env}(a)).
Once it pressed against the obstacle (Fig.~\ref{fig_exp_simple_env}(b)), the factor bounded the robot on one side and, compared with the estimate without the environment, reduced the $3\sigma$ tip uncertainty from $12.6$ to $5.9$~mm along $x$ and from $16.6$ to $9.8$~mm along the optical axis, averaged over the contact (Supplementary Video S1).
During contact, the tip error had an RMS of $2.0$~mm, compared with $4.2$~mm without the environment, and every sample remained within the $3\sigma$ bounds.

\begin{figure}[t!]
\centerline{\includegraphics[width=0.75\linewidth]{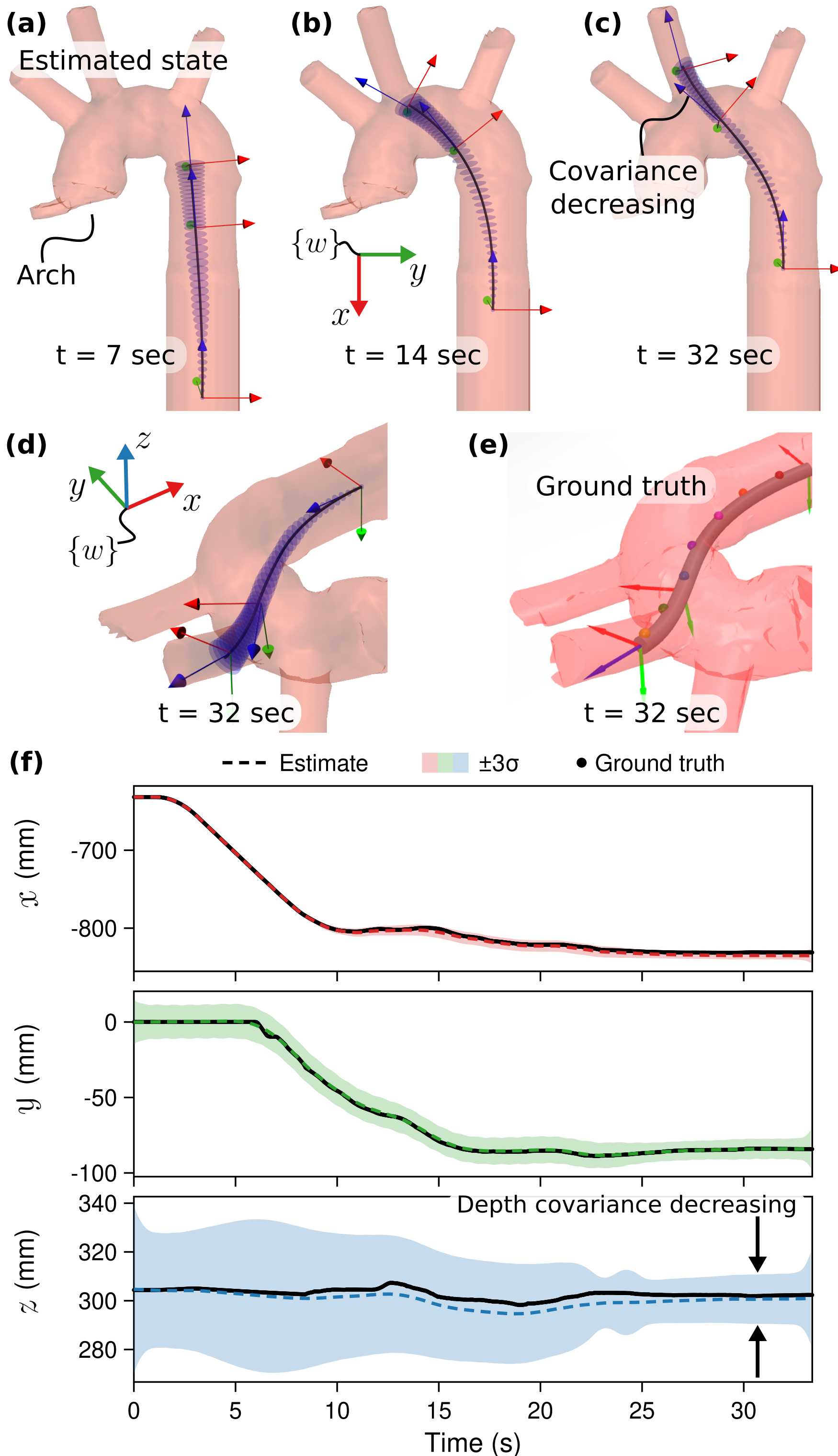}}
\vspace{-8pt}
\caption{
Simulated navigation from the descending aorta into a supra-aortic branch, estimated from the frontal view with the environment factor (case C of Table~\ref{tab_arch}).
(a)--(c)~Estimate at $t=7$, $14$, and $32$~s, with backbone frames and position covariance ellipsoids (purple).
(d)~Oblique view of the estimate at $t=32$~s and (e)~ground truth.
(f)~Estimated tip position in $\{w\}$ (dashed) with $\pm3\sigma$ bounds (shaded) and ground truth (black).
The known vessel geometry let the estimator track the tip through the arch from one view.
}
\label{fig_exp_SIM_arch}  
\end{figure}

\begin{table}[t]
\caption{Mean Tip RMSE (\% of $L$) and NEES in the Aortic Arch}
\label{tab_arch}
\centering
\setlength{\tabcolsep}{4pt}
\begin{tabular}{@{}cccrccrcc@{}}
\toprule
 & & & \multicolumn{3}{c}{Simulation} & \multicolumn{3}{c}{Hardware} \\
\cmidrule(lr){4-6} \cmidrule(l){7-9}
Case & Views & Env. & Images & RMSE & NEES & Images & RMSE & NEES \\
\midrule
A & 1 &            & 42  & 2.5 & 2.8 & 42 & 20.0 & 6.9 \\
B & 3 &            & 126 & 0.9 & 1.5 & 85 & 10.5 & 5.0 \\
C & 1 & \checkmark & 42  & 1.6 & 2.6 & 42 & 11.3 & 5.8 \\
D & 3 & \checkmark & 126 & 0.9 & 1.6 & 85 & 8.9  & 5.0 \\
\bottomrule
\end{tabular}
\end{table}

\subsubsection{Aortic Arch}
Finally, an operator watching the rendered scene teleoperated the robot from the descending aorta into a target branch in three trials.
The camera cycled through the frontal, LAO, and RAO views available during mechanical thrombectomy, taking an image every $0.26$~s.
Each trial of $31$ to $33$~s took $5$ to $21$~s to estimate both the state mean and covariance.
Because MuJoCo collides only convex geometry, the scanned aortic arch wall was decomposed into convex parts with CoACD~\cite{wei2022coacd} for improved contact modeling.
Each trial was estimated from the frontal images alone or from all three views, with and without the environment factor (Table~\ref{tab_arch}).
With the frontal view alone, the environment factor lowered the tip error from $2.5\%$ to $1.6\%$ of the robot length and tracked the tip through the arch (Fig.~\ref{fig_exp_SIM_arch}).
Adding the LAO and RAO views lowered it to $0.9\%$ with or without the environment factor, since with three views the estimate seldom reached the vessel wall where the factor acts.
The tip NEES remained between $1.5$ and $2.8$ in every case, so the estimate was consistent.

\begin{figure}[t!]
\centerline{\includegraphics[width=0.8\linewidth]{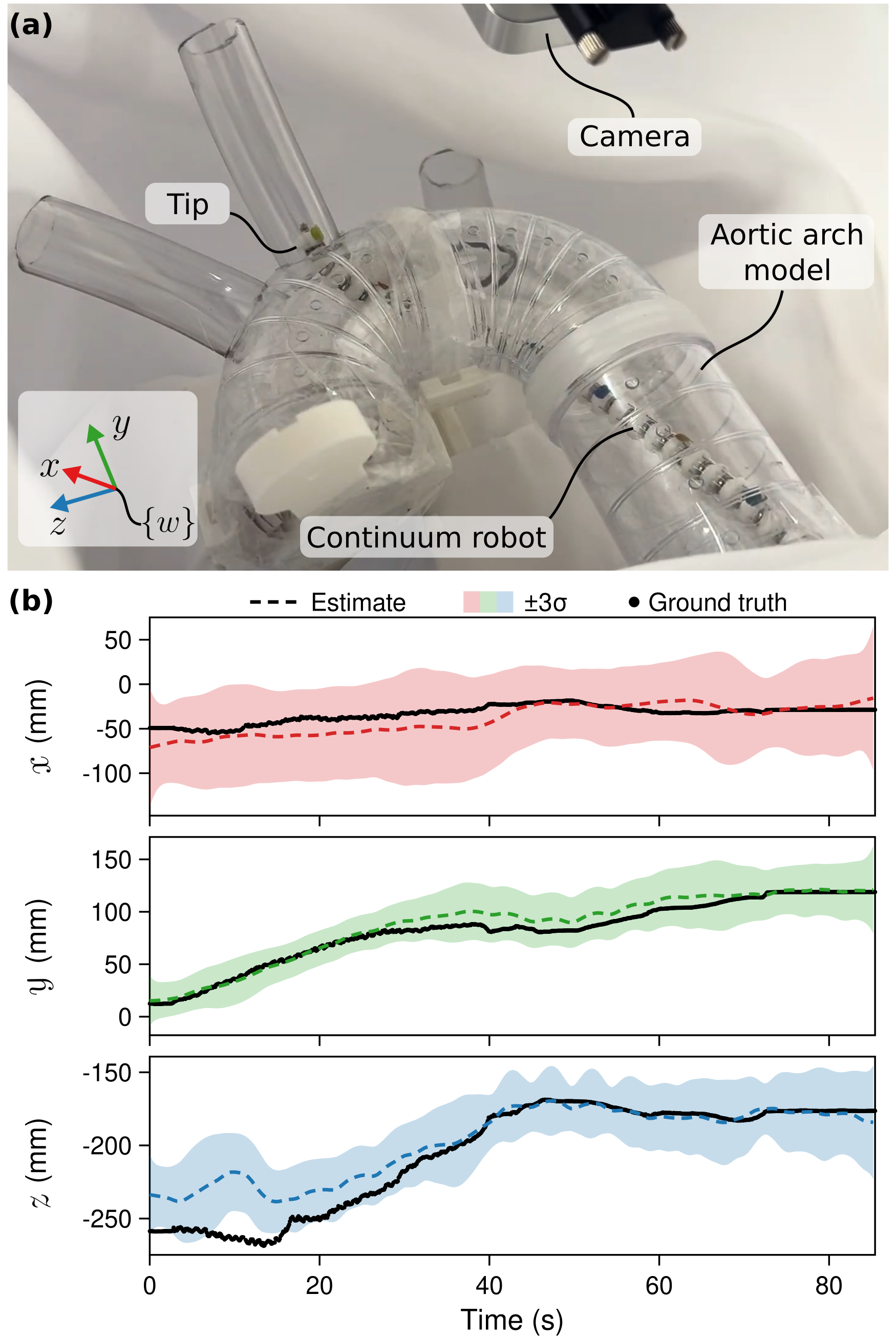}}
\vspace{-8pt}
\caption{
Estimation on hardware in an aortic arch phantom.
(a)~Robot teleoperated into the target branch of the transparent phantom, imaged by a camera on a serial robot arm from three views.
(b)~Estimated tip position (dashed) with $\pm3\sigma$ bounds (shaded) and ground truth (black) over one trial, from all three views with the environment factor (case D of Table~\ref{tab_arch}).
The estimate tracked the tip within its bounds for most of the trial.
}
\label{fig_exp_HW_arch}  
\end{figure}

\subsection{Hardware} \label{sec_exp_hw}

We next demonstrated the framework on hardware.
A linear stage inserted a two-segment tendon-driven continuum robot~\cite{shentu2026rigid} ($L=197.7$~mm, radius $5$~mm, four tendons per segment) into the scanned aortic arch phantom (Fig.~\ref{fig_exp_HW_arch}).
The robot consisted of 3D printed disks and laser cut metal saddle joints along a hollow spring backbone.
An Intel RealSense D405 camera on a Franka Panda robot arm cycled through the frontal, LAO, frontal, and RAO views and took a $1280\times720$ image about every $1.2$~s, in which it looked for six colored markers on the robot.
An electromagnetic (EM) sensor (Northern Digital Inc.) provided the camera pose and, for evaluation only, the ground truth at three sample locations along the robot, namely the base, the end of the proximal segment, and the tip.

In three trials, an operator viewing the phantom directly teleoperated the robot from the arch into the target branch, and each trial of $85$ to $104$~s took $12$ to $23$~s to estimate the state mean and covariance.
With all three views and the environment factor, the tip error over each trial averaged $17.5$~mm, or $8.9\%$ of the robot length, and the tip NEES averaged $5.0$ (Table~\ref{tab_arch}).
The NEES above three arose while the robot traversed the arch and pressed on its wall.
The main source of error was the inability to detect markers.
Glare and the limited transparency of the phantom hid some markers, which would not be a limitation under  fluoroscopy.
On average only $3.3$ of the $6$ markers were detected in each image, compared with $6.7$ of $7$ in simulation.
With the frontal view alone, the environment factor lowered the tip error from $20.0\%$ to $11.3\%$, close to the $10.5\%$ of three views without it, so using the known environment compensated for sparse and incomplete imaging.

\section{Conclusion}
In this work, we presented a probabilistic state estimator for multi-segment continuum robots that fuses planar images, commanded actuation, and a known environment in a single factor graph over arclength and time, without sensors on the robot.
Contact with the known environment reduced the uncertainty of the estimate, and in a simulated aortic arch imaged from a single view, the environment factor lowered the tip error from $2.5\%$ to $1.6\%$ of the robot length.
On hardware, the estimator tracked the robot into the target branch of an aortic arch phantom, and with a single view the environment factor lowered the tip error from $20.0\%$ to $11.3\%$ of the robot length.
The estimator assumed a static, rigid environment. 
This approximation suits the aortic arch, which is anchored in the mediastinum, and rigid structures such as pipes in industrial inspection, but not environments that move or deform substantially.
Future work will model environment motion and deformation, estimate the registration of the environment jointly with the robot state, and use a clinical C-arm with radiopaque markers on a continuum robot at clinical scale.
The continuous space-time estimate could also inform control, and imaging poses could be selected actively to reduce uncertainty while limiting radiation exposure.
Overall, these results showed that the anatomy known before a procedure and the commands sent to the robot can take the place of embedded sensors, so that sparse planar images were enough to estimate the full shape of a continuum robot together with its uncertainty.
This brings sensorless continuum robots closer to safe, image-guided navigation.


\bibliography{biblio}

\end{document}